\pdfoutput=1
\documentclass[11pt]{article}

\usepackage[]{EMNLP2023}

\usepackage{hyperref}
\usepackage{url}

\usepackage{times}
\usepackage{latexsym}
\usepackage{booktabs}
\usepackage{graphicx}
\usepackage[T1]{fontenc}
\usepackage{makecell}
\usepackage[utf8]{inputenc}

\usepackage{microtype}

\usepackage{inconsolata}

\usepackage{amsthm,amssymb,amsfonts,amsmath}

\usepackage{caption,subfigure,graphicx,epstopdf,multirow,multicol,booktabs,verbatim,wrapfig,bm}

\usepackage{xcolor}

\definecolor{BrickRed}{RGB}{178, 34, 34}
\definecolor{MidnightBlue}{RGB}{0, 0, 128}

\newcommand{\redbf}[1]{\bf{\textcolor{BrickRed}{#1}}}
\newcommand{\bluebf}[1]{\bf{\textcolor{MidnightBlue}{#1}}}

\newcommand{\eg}{\textit{e.g., }}

\newcommand{\wrt}{\textit{w.r.t. }}

\title{Can Large Language Models Empower Molecular Property Prediction?}

\author{
Chen Qian\textsuperscript{\rm 1}, 
Huayi Tang\textsuperscript{\rm 1},
Zhirui Yang\textsuperscript{\rm 1},
Hong Liang\textsuperscript{\rm 2},
Yong Liu\textsuperscript{\rm 1 \thanks{\quad Corresponding author.}}  \\
\textsuperscript{\rm 1} Renmin University of China  
\textsuperscript{\rm 2} Peking University \\
\texttt{\{qianchen2022,huayitang,yangzhirui,liuyonggsai\}@ruc.edu.cn}, lho@stu.pku.edu.cn \\
}

\begin{document}
\maketitle
\begin{abstract}
Molecular property prediction has gained significant attention due to its transformative potential in multiple scientific disciplines. Conventionally, a molecule graph can be represented either as a graph-structured data  or a SMILES text.
Recently, the rapid development of Large Language Models (LLMs) has revolutionized the field of NLP. 
Although it is natural to utilize LLMs to assist in understanding molecules represented by SMILES, the exploration of how LLMs will impact molecular property prediction is still in its early stage. 
In this work, we advance towards this objective through two perspectives: zero/few-shot molecular classification, and using the new explanations generated by LLMs as representations of molecules.
To be specific, we first prompt LLMs to do in-context molecular classification and evaluate their performance. After that, we employ LLMs to generate semantically enriched explanations for the original SMILES and then leverage that to fine-tune a small-scale LM model for multiple downstream tasks. 
The experimental results highlight the superiority of text explanations as molecular representations across multiple benchmark datasets, and confirm the immense potential of LLMs in molecular property prediction tasks.
Codes are available at \url{https://github.com/ChnQ/LLM4Mol}.
\end{abstract}

\section{Introduction}
As a cutting-edge research topic at the intersection of artificial intelligence and chemistry, molecular property prediction has drawn increasing interest due to its transformative potential in multiple scientific disciplines such as virtual screening , drug design and discovery \cite{zheng2019onionnet, maia2020structure, gentile2022artificial}, to name a few. 
Based on this, the effective modeling of molecular data constitutes a crucial prerequisite for AI-driven molecular property prediction tasks \cite{rong2020self, wang2022chemicalreactionaware}.
In the previous literature, on one hand, molecules can be naturally represented as graphs with atoms as nodes and chemical bonds as edges. Therefore, Graph Neural Networks (GNNs) can be employed to handle the molecular data \cite{kipf2016semi, xu2018powerful, sun2019infograph, rong2020self}. 
Simultaneously, the other line of research explores the utilization of NLP-like techniques to process molecular data \cite{wang2019smiles, honda2019smiles, wang2022chemicalreactionaware}, since in many chemical databases \cite{irwin2005zinc, gaulton2017chembl}, molecular data is commonly stored as SMILES (Simplified Molecular-Input Line-Entry System) \cite{weininger1988smiles} strings, a textual representation of molecular structure following strict rules. 
% Fig1
\begin{figure}[t]
  \centering
  \includegraphics[width=\linewidth]{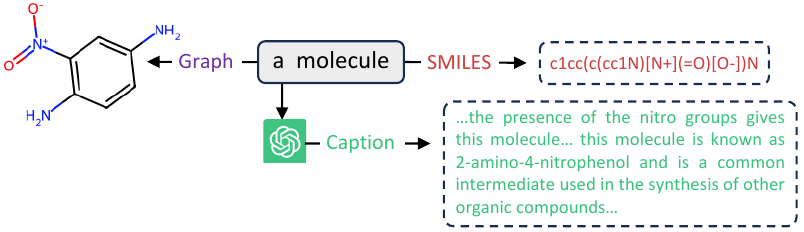}
  \caption{Different representation paradigms for a molecule.}
  \label{fig-intro}
\end{figure}

% overview
\begin{figure*}[t]
    \centering
    \includegraphics[width=\linewidth]{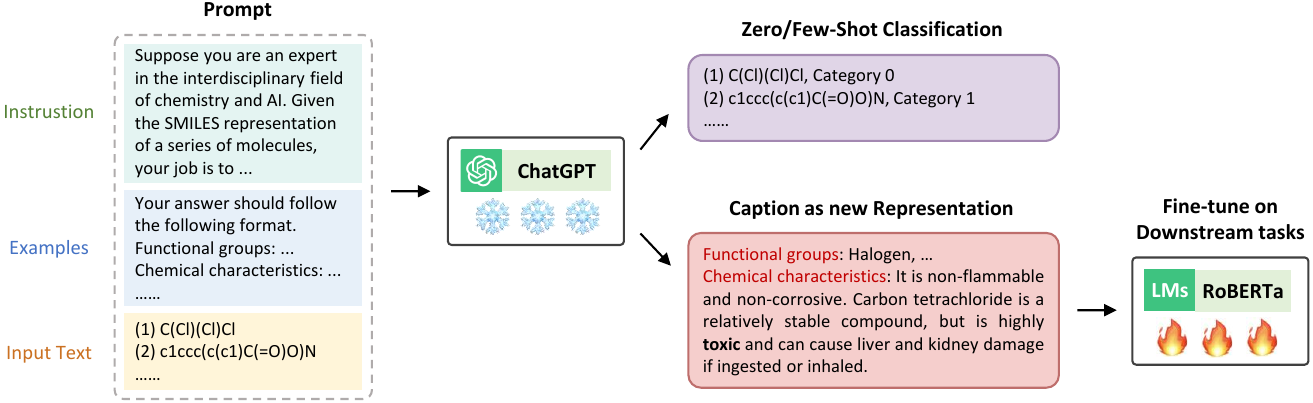}
    \caption{Overview of LLM4Mol.}
    \label{fig-overview}
    \vspace{-5mm}
\end{figure*}

In recent years, the rapid development of LLMs have sparked a paradigm shift and opened up unprecedented opportunities in the field of NLP \cite{zhao2023survey, zhou2023comprehensive}. 
Those models demonstrate tremendous potential in addressing various NLP tasks and show surprising abilities (i.e., emergent abilities \cite{wei2022emergent}).
Notably, ChatGPT~\cite{OpenAI2023GPT4TR} is the state-of-the-art AI conversational system developed by OpenAI in 2022, which possesses powerful text understanding capabilities and has been widely applied across various vertical domains.

Note that, since molecules can be represented as SMILES sequences, it is natural and intuitive to employ LLMs with rich world knowledge to handle molecular data.
For instance, as depicted in Figure~\ref{fig-intro}, given the SMILES line of a molecule, ChatGPT can accurately describe the functional groups, chemical properties, and potential pharmaceutical applications \wrt the given molecule. We believe that such textual descriptions are meaningful for assisting in molecular-related tasks.

However, the application of LLMs in molecular property prediction tasks is still in its primary stages. 
In this paper, we move towards this goal from two perspectives: zero/few-shot molecular classification task, and generating new explanations for molecules with original SMILES.
Concretely, inspired by the astonishing in-context learning capabilities \cite{brown2020language} of LLMs, we first prompt ChatGPT to perform in-context molecular classification.
Then, we propose a novel molecular representation called \underline{C}aptions \underline{a}s new \underline{R}epresentation (CaR), which leverages ChatGPT to generate informative and professional textual analyses for SMILES. Then the textual explanation can serving as new representation for molecules, as illustrated in Figure~\ref{fig-intro}.
Comprehensive experimental results highlight the remarkable capabilities and tremendous potential of LLMs in molecular property prediction tasks.
We hope this work could shed new insights in model design of molecular property prediction tasks enpowered by LLMs.

\section{Method}
In this section, we will elaborate on our preliminary exploration of how LLMs can serve molecular property prediction tasks. 
% through two approaches: in-context learning and CaR.

\noindent
\textbf{Zero/Few-shot Classification.}
With the continuous advancement of LLMs, In-Context Learning (ICL) \cite{brown2020language} has emerged as a new paradigm for NLP. 
Using a demonstration context that includes several examples written in natural language templates as input, LLMs can make predictions for unseen input without additional parameter updates \cite{liu2022close, lu2021fantastically, wu2022self, wei2022chain}.
Therefore, we attempt to leverage the ICL capability of ChatGPT to assist in molecular classification task by well-designed prompts, as shown in Figure~\ref{fig-overview}. This paradigm makes it much easier to incorporate human knowledge into LLMs by changing the demonstration and templates.

\noindent
\textbf{Captions as New Representations.} With vivid world knowledge and amazing reasoning ability, LLMs have been widely applied in various AI domains \cite{he2023explanations, liu2023chatgpt}. Also, we reckon that LLMs can empower LLMs can greatly contribute to the understanding of molecular properties. Taking a commonly used dataset in the field of molecular prediction for a toy example, PTC \cite{helma2001predictive} is a collection of chemical molecules that reports their carcinogenicity in rodents.
We conduct a keyword search using terms such as `toxicity' `cancer', and `harmful' to retrieve all explanations generated by ChatGPT for the originally SMILES-format PTC dataset. 
Interestingly, we observed that the majority of these keywords predominantly appeared in entries labeled as $\text{-}1$. 
This demonstrates that ChatGPT is capable of providing meaningful and distinctive professional explanations for the raw SMILES strings, thereby benefiting downstream tasks.

Towards this end, we propose to leverage ChatGPT to understand the raw SMILES strings and generate textual descriptions that encompass various aspects such as functional groups, chemical properties, pharmaceutical applications, and beyond. 
Then, we fine-tune a pre-trained small-scale LM (\eg RoBERTa \cite{liu2020roberta}) on various downstream tasks, such as molecular classification and properties prediction. 

\begin{table*}[t!]
\centering
\caption{Testing evaluation results on several benchmark datasets with \textbf{Random Splitting}. For classification task reporting ACC and ROC-AUC (\%, mean $\pm$ std), for regression tasks reporting RMSE (mean $\pm$ std). $\uparrow$ for higher is better, $\downarrow$ contrarily. $\ddagger$ denotes the results cited from origin paper. CoR with superior result is \redbf{highlighted}.}
\label{tb:main_result}
\resizebox{1.0\textwidth}{!}{
\setlength{\tabcolsep}{1.5mm}{
\begin{tabular}{@{}ll|ccc|cc|cc@{}}
\toprule
  & \multirow{2}{*}{Method} & \multicolumn{3}{c|}{\makecell{ACC $\uparrow$}} &  \multicolumn{2}{c|}{ROC-AUC $\uparrow$} & \multicolumn{2}{c}{RMSE $\downarrow$} \\
  &  & MUTAG & PTC & AIDS & Sider & ClinTox & Esol & Lipo \\ 
   \hline
\hline\specialrule{0em}{2pt}{2pt}
\multirow{5}{*}{\rotatebox{90}{GNNs}} & GCN  &  $90.00 \pm 4.97$  & $62.57 \pm 4.13$ & $78.68 \pm 3.36$ & $64.24 \pm 5.61$ & $91.88 \pm 1.45$ & $0.77 \pm 0.05$ & $0.80 \pm 0.04$ \\ \cmidrule(l){2-9} 
              & GIN  &  $89.47 \pm 4.71$  & $58.29 \pm 5.88$ & $78.01 \pm 1.77$ & $66.19 \pm 5.10$ & $92.08 \pm 1.11$ & $0.67 \pm 0.04$ & $0.79 \pm 0.03$ \\ \cmidrule(l){2-9} 
              & ChebyNet & $64.21 \pm 5.16$  &  $61.43 \pm 4.29$  & $79.74 \pm 1.78$ & $80.68 \pm 5.10$  & $91.48 \pm 1.50$ & $0.75 \pm 0.04$ & $0.85 \pm 0.04$ \\ \cmidrule(l){2-9} 
              & D-MPNN$^\ddagger$ & - & - & - & $66.40 \pm 2.10$ & $90.60 \pm 4.30$ & $0.58 \pm 0.05$ & $0.55 \pm 0.07$ \\ \hline

\hline\specialrule{0em}{2pt}{2pt}
\multirow{4}{*}{\rotatebox{90}{SMILEs}} & ECFP4-MLP  &  $96.84 \pm 3.49$  & $85.71 \pm 7.67$ & $94.64 \pm 3.14$ & $90.19 \pm 4.88$ & $95.81 \pm 2.09$  & $0.60 \pm 0.11$ & $0.60 \pm 0.16$ \\ \cmidrule(l){2-9} 
              & SMILES-Transformer$^\ddagger$  & -  & - & - & - &  $95.40$ & $0.72$ & $0.92$ \\ \cmidrule(l){2-9} 
              & MolR$^\ddagger$  & - & - & - & - & $91.60 \pm 3.90$ & - & - \\ \hline
\hline\specialrule{0em}{2pt}{2pt}
\multirow{3}{*}{\rotatebox{90}{LLM}}
& \textbf{CaR}$_{Roberta}$   &  $91.05 \pm 3.37$    & $\redbf{93.14 \pm 3.43}$ & $94.37 \pm 1.19$ & $88.81 \pm 2.65$ & $\redbf{99.80 \pm 0.43}$  & $\redbf{0.45 \pm 0.04}$ & $\redbf{0.47 \pm 0.03}$ \\
& $\Delta_{GNNs}$   &   $\redbf{+12\%}$   & $\redbf{+53\%}$ & $\redbf{+20\%}$ & $\redbf{+30\%}$ & $\redbf{+9\%}$ & $\redbf{-35\%}$ & $\redbf{-37\%}$ \\  
& $\Delta_{NLP}$   &   $\bluebf{-6\%}$   & $\redbf{+9\%}$ & $\bf{+0\%}$ & $\bluebf{-2\%} $ & $\redbf{+6\%}$ & $\redbf{-32\%}$ & $\redbf{-38\%}$\\   
\bottomrule
\end{tabular}}}
\vspace{-3mm}
\end{table*}

\section{Experiments}

\subsection{Setup}
\noindent
\textbf{Datasets. }
To comprehensively evaluate the performance of CaR, we conduct experiments on 9 datasets spanning molecular classification tasks and molecular regression tasks.  
\romannumeral1) 3 classification datasets from TUDataset \cite{Morris+2020}: MUTAG, PTC, AIDS.
\romannumeral2) 4 classification datasets from MoleculeNet \cite{wu2018moleculenet}: Sider, ClinTox, Bace, BBBP.
\romannumeral3) 2 regression datasets from MoleculeNet: Esol, Lipophilicity.

\noindent
\textbf{Baselines.}
We compare CaR with the following baselines:
\romannumeral1) GNN-based methods, GCN \cite{kipf2016semi}, GIN \cite{xu2018powerful}, ChebyNet \cite{defferrard2016convolutional}, D-MPNN \cite{yang2019analyzing}, GraphMVP \cite{liu2022pretraining}, InfoGraph \cite{sun2019infograph}, G-Motif \cite{rong2020self}, Mole-BERT \cite{xia2023molebert}.
\romannumeral2) SMILES-based methods, ECFP \cite{rogers2010extended}, SMILES-Transfor \cite{honda2019smiles}, MolR \cite{wang2022chemicalreactionaware}, ChemBERTa \cite{chithrananda2020chemberta}, MolKD \cite{zeng2023molkd}.

% few-shot exp
\begin{figure}[t]
    \centering
    \includegraphics[width=\linewidth]{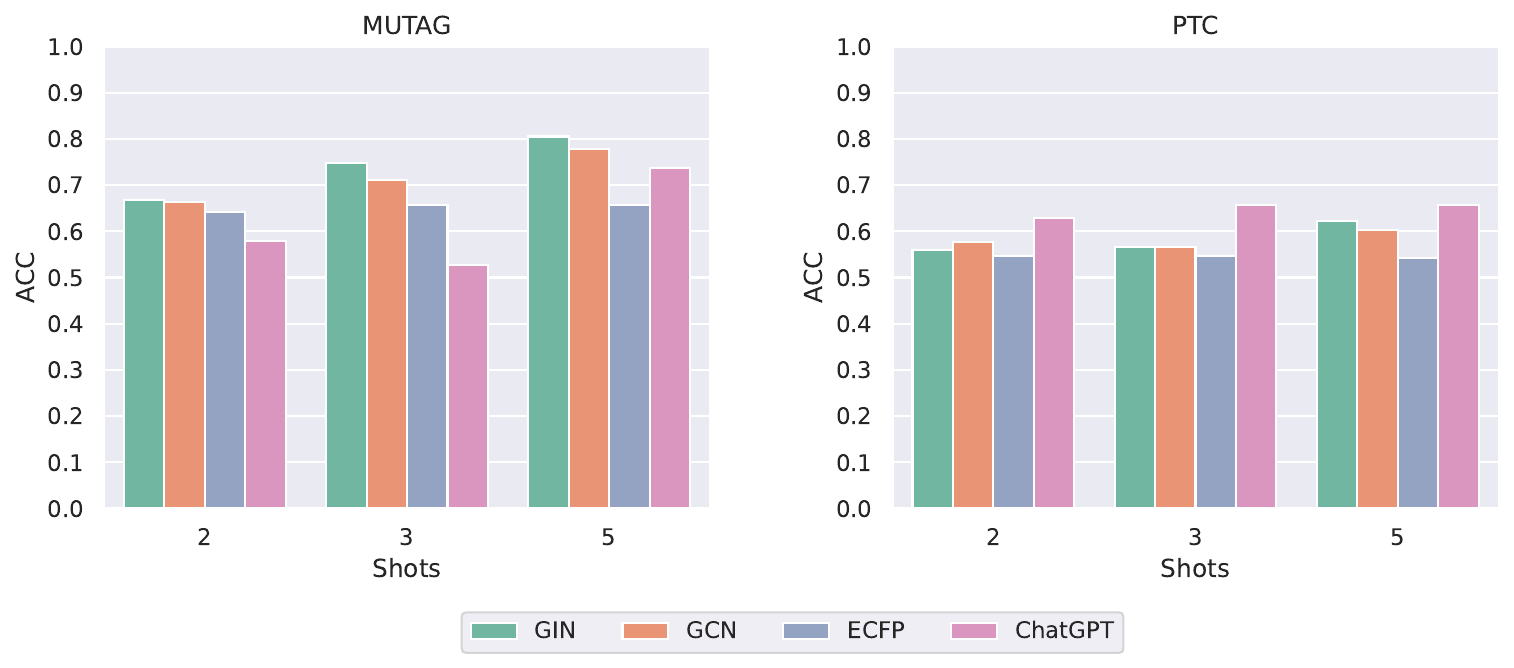}
    \vspace{-6mm}
    \caption{Few-shot classification results on MUTAG and PTC by classical models and ChatGPT.}
    \label{fig:few-bar}
    \vspace{-6mm}
\end{figure}

\noindent
\textbf{Settings. } 
For all datasets, we perform a $8/1/1$ splitting for train/validate/test, where the best average performance (and standard variance) on the test fold is reported.
Specially, we perform a $10$-fold cross-validation (CV) with a holdout fixed test for random split datasets; conduct experiments for scaffold splitting datasets with $5$ random seeds.
Small-scale LMs are implemented using the Hugging Face transformers library \cite{wolf-etal-2020-transformers} with default parameters.
% For more experimental details please refer to Appendix.xx, 

\begin{table*}[t!]
\centering
\caption{Testing evaluation results of different methods on benchmark datasets with \textbf{Scaffold Splitting}. The remaining settings keep consistent with Table~\ref{tb:main_result}.}
\label{tb:main_result2}
\resizebox{0.91\textwidth}{!}{
\setlength{\tabcolsep}{2.5mm}{
\begin{tabular}{@{}ll|cccc|cc@{}}
\toprule
    & \multirow{2}{*}{Method}  &  \multicolumn{4}{c|}{\makecell{ROC-AUC $\uparrow$}} & \multicolumn{2}{c}{RMSE $\downarrow$} \\
    &  & Sider & ClinTox & Bace & BBBP & Esol & Lipo \\ 
                       \hline
\hline\specialrule{0em}{2pt}{2pt}
\multirow{8}{*}{\rotatebox{90}{GNNs}} 
& GCN  & $55.81 \pm 2.92$ &  $50.32 \pm 2.46$ & $76.78 \pm 4.74$ & $71.90 \pm 5.35$ & $1.09 \pm 0.11$ & $0.88 \pm 0.03$  \\ \cmidrule(l){2-8} 
& GIN  & $58.86 \pm 2.57$ &  $51.79 \pm 5.18$ & $77.05 \pm 5.68$ & $75.30 \pm 4.66$ & $1.26 \pm 0.49$ & $0.88 \pm 0.02$ \\ \cmidrule(l){2-8} 
& ChebyNet & $60.87 \pm 1.68$ &  $52.92 \pm 9.36$  & $77.31 \pm 3.55$ & $73.89 \pm 4.95$ & $1.09 \pm 0.08$ & $0.89 \pm 0.04$ \\ \cmidrule(l){2-8} 
& InfoGraph$^\ddagger$ & $59.20 \pm 0.20$ & $75.10 \pm 5.00$ & $73.90 \pm 2.50$ & $69.20 \pm 0.80$ & - & -  \\ \cmidrule(l){2-8} 
& G-Motif$^\ddagger$ & $60.60 \pm 1.10$ & $77.80 \pm 2.00$ & $73.40 \pm 4.00$ & $66.40 \pm 3.40$ & - & - \\ \cmidrule(l){2-8} 
& GraphMVP-C$^\ddagger$ & $63.90 \pm 1.20$ & $77.50 \pm 4.20$ & $81.20 \pm 0.90$ & $72.40 \pm 1.60$ & $1.03$ & $0.68$ \\ \cmidrule(l){2-8}
& Mole-BERT$^\ddagger$ & $62.80 \pm 1.10$ & $78.90 \pm 3.00$ & $80.80 \pm 1.40$ & $71.90 \pm 1.60$ & $1.02 \pm 0.03$ & $0.68 \pm 0.02$ \\ \hline

\hline\specialrule{0em}{2pt}{2pt}
\multirow{4}{*}{\rotatebox{90}{SMILEs}} 
& ECFP4-MLP & $64.86 \pm 3.45$ & $52.93 \pm 5.92$ & $81.58 \pm 4.02$ & $73.37 \pm 6.05$ & $1.77 \pm 0.25$ & $1.03 \pm 0.04$ \\ \cmidrule(l){2-8} 
& ChemBERTa$^\ddagger$  & - &  $73.30$ & - & $64.30$ & - & - \\ \cmidrule(l){2-8} 
& MolKD$^\ddagger$  & $61.30 \pm 1.20$ &  $83.80 \pm 3.10$ & $80.10 \pm 0.80$ & $74.80 \pm 2.30$ & - & - \\ \hline
\hline\specialrule{0em}{2pt}{2pt}
\multirow{3}{*}{\rotatebox{90}{LLM}}
& \textbf{CaR}$_{Roberta}$  & $58.06 \pm 1.80$ & $\redbf{84.16 \pm 17.63}$  & $80.73 \pm 1.42$ & $\redbf{81.99 \pm 4.19}$ & $\redbf{0.96 \pm 0.09}$ & $1.02 \pm 0.06$ \\
& $\Delta_{GNNs}$  & $\bluebf{-3\%}$ & $\redbf{+30\%}$ & $\redbf{+5\%}$ & $\redbf{+15\%}$  & $\redbf{-13\%}$ & $\bluebf{+27\%}$ \\
& $\Delta_{NLP}$  & $\bluebf{-9\%}$ & $\redbf{+22\%}$ & $\bluebf{-1\%}$ & $\redbf{+19\%}$ & $\redbf{-46\%}$ & $\redbf{-1\%}$ \\  
\bottomrule
\end{tabular}}}
\vspace{-1mm}
\end{table*}

\begin{figure}[t]
    \centering
    \includegraphics[width=0.7\linewidth]{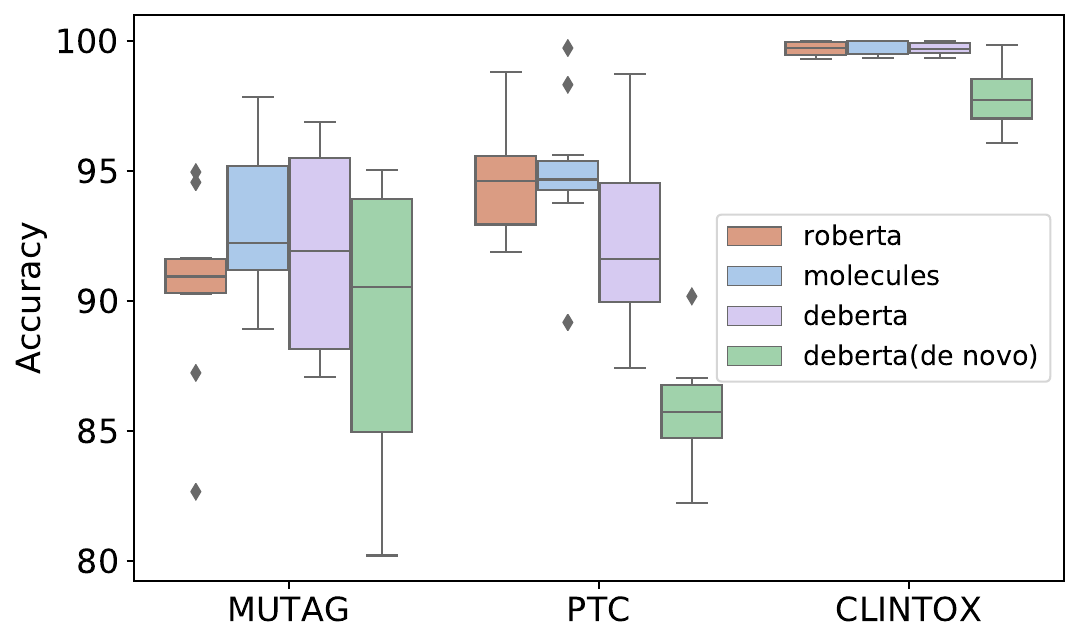}
    \vspace{-2mm}
    \caption{Performance of CaR by replacing Small LMs.}
    \label{fig:ablation}
    \vspace{-3mm}
\end{figure}

\begin{figure*}[t]
    \subfigure{
       \centering
        \includegraphics[width=0.33\linewidth]{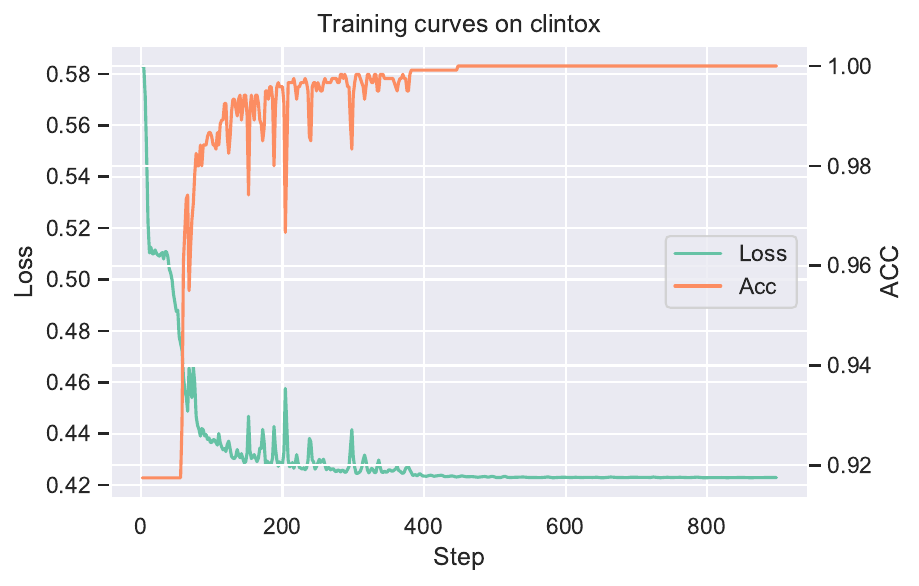}
    }\subfigure{
        \centering
        \includegraphics[width=0.33\linewidth]{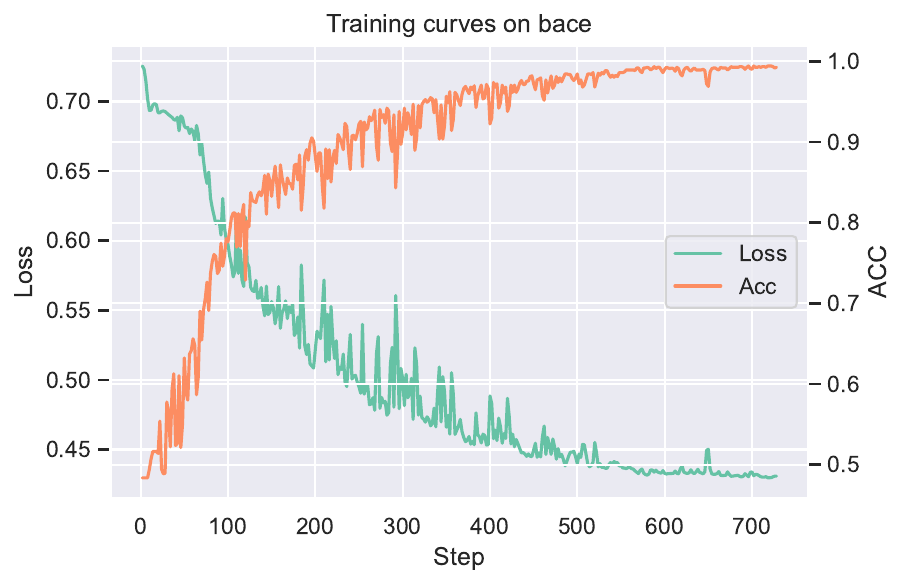}
    }\subfigure{
        \centering
        \includegraphics[width=0.33\linewidth]{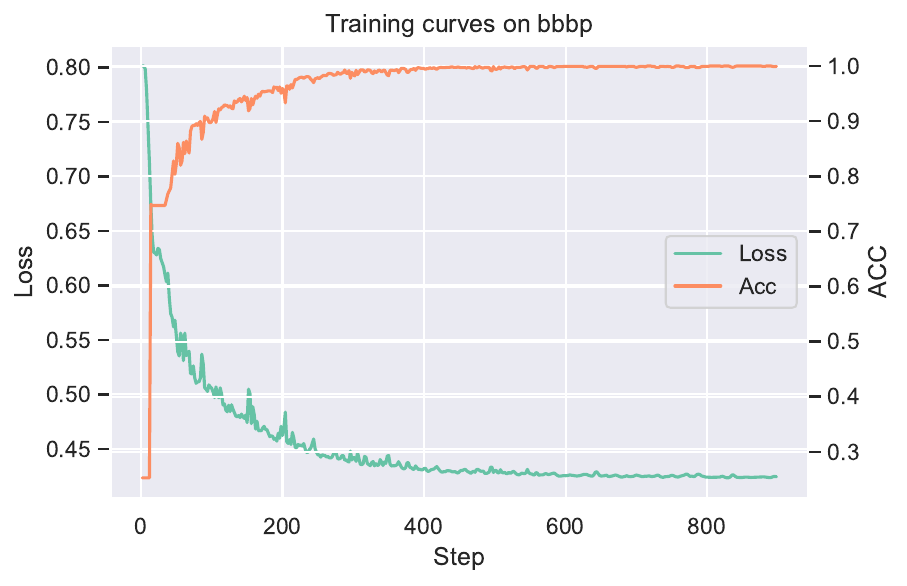}
    }
    \vspace{-8mm}
    \caption{ The loss value (Loss) and accuracy value (ACC) during training process.}
    \vspace{-3mm}
    \label{fig:shoulian}
\end{figure*}

\subsection{Main Results}
\noindent
\textbf{How does ChatGPT perform on zero/few-shot molecular classification?} Figure~\ref{fig:few-bar} illustrates the few-shot learning capabilities of ChatGPT, traditional GNNs, and ECFP on two datasets. 
It is observed that ChatGPT underperforms compared to traditional methods for MUTAG, whereas conversely for PTC. Furthermore, see Figure~\ref{fig:few-line}, as the number of shots increases, ChatGPT demonstrates an upward trend in performance for both datasets. 
These results indicate that ChatGPT possesses a certain level of few-shot molecular classification capability. 
However, throughout the experiments, we find that ChatGPT's classification performance was not consistent for the same prompt, and different prompts also have a significant impact on the results. Therefore, it is crucial to design effective prompts that incorporate rational prior information to achieve better zero/few-shot classification.

\noindent
\textbf{How does CaR perform compared with existing methods on common benchmarks?}
The main results for comparing the performance of different methods on several benchmark datasets are shown in Table~\ref{tb:main_result} and Table~\ref{tb:main_result2}. From the tables, we obtain the following observation: 
% random
\romannumeral1) Under the random split setting, CaR achieves superior results on almost all datasets, whether in classification or regression tasks. Remarkably, CaR exhibits a significant performance improvement of $53\%$ compared to traditional methods on the PTC dataset.
% scaff
\romannumeral2) For Scaffold splitting, one can observe that compared to other models, LLM demonstrates comparable results on Sider and Bace with slightly less superior; in the Lipo regression task, CaR falls short compared to GNNs; However, CoR achieves notable performance improvements on the remaining datasets.
These observations indicate LLMs' effectiveness and potential in enhancing molecular predictions across various domains.

\noindent
\textbf{Convergence Analysis.} In Figure \ref{fig:shoulian}, we plot the ROC-AUC and loss curves on three datasets to verify CaR's convergence. 
One can observe that the loss value decreases rapidly in the first several steps and then continuously decrease in a fluctuation way until convergence. Also, the ROC-AUC curve exhibits an inverse and corresponding trend. These results demonstrate the convergence of CaR.

\noindent
\textbf{Replace Small-scale LMs.} 
To validate the effectiveness of CaR, we further fine-tune two additional pre-trained LMs (DeBERTa \cite{he2021deberta}, adaptive-lm-molecules \cite{blanchard2023adaptive}) and also train a non-pretrained DeBERTa from scratch. The results are plotted in Figure~\ref{fig:ablation}. One can observe that different pre-trained LMs exhibit similar performance, and generally outperform the LM trained from scratch, which validate the effectiveness of CaR.

\section{Conclusion}
In this work, we explore how LLMs can contribute to molecular property prediction from two perspectives, in-context classification and generating new representation for molecules.
This preliminary attempt highlights the immense potential of LLM in handling molecular data. In future work, we attempt to focus on more complex molecular downstream tasks, such as generation tasks and 3D antibody binding tasks. 
% By doing so, we can further explore the capabilities of LLM and expand its applications in diverse areas of molecular research.

\section*{Limitations}

\noindent
\textbf{Lack of Diverse LLMs.} In this work, we primarily utilized ChatGPT as a representative of LLMs. However, the performance of other LLMs on molecular data has yet to be explored, such as the more powerful GPT-4 \cite{OpenAI2023GPT4TR} or domain-specific models like MolReGPT \cite{li2023empowering}.

\noindent
\textbf{Insufficient Mining of Graph Structures.} While we currently model molecular prediction tasks solely as NLP tasks, we acknowledge the crucial importance of the graph structure inherent in molecules for predicting molecular properties. How to further enhance the performance of our framework by mining graph structured information is worth exploring.

\noindent
\textbf{Beyond SMILES.} In this work, we focus on small molecule data that can be represented as SMILES strings. However, in practical biochemistry domains, there is a wide range of data, such as proteins, antibodies, and other large molecules, that cannot be represented using SMILES strings. Therefore, the design of reasonable sequential representations for the large molecules with 3D structure to LLMs of is an important and urgent research direction to be addressed.

\bibliography{anthology,custom}
\bibliographystyle{acl_natbib}

\appendix

\newpage

\section{N-shot Results}
\begin{figure}[h]
    \centering
    \includegraphics[width=\linewidth]{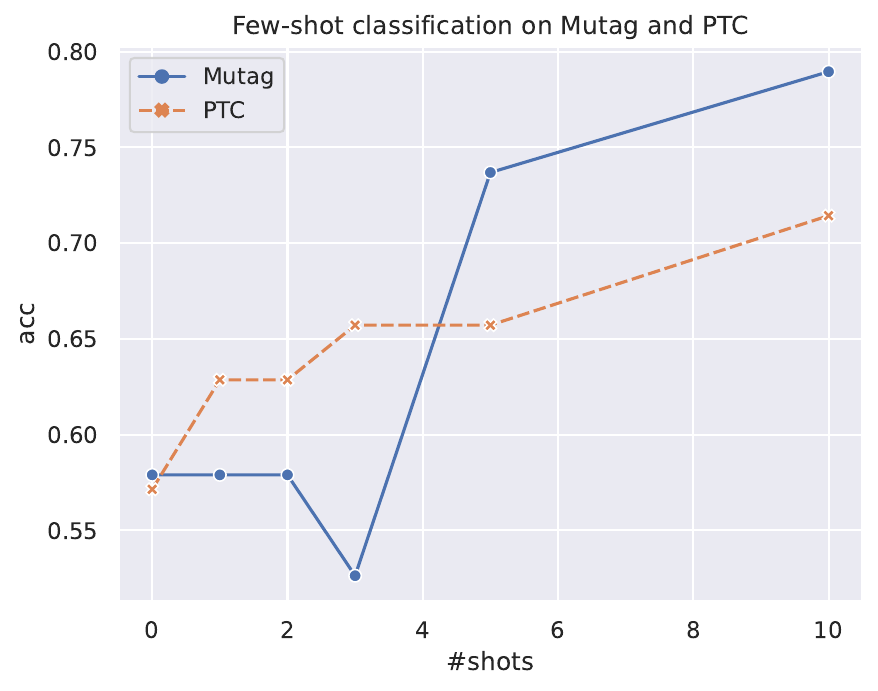}
    \caption{The impact of \#Shots on Few-shot classification on MUTAG and PTC by ChatGPT.}
    \label{fig:few-line}
\end{figure}

\end{document}